\documentclass[10pt,twocolumn,letterpaper]{article}

\usepackage{iccv}
\usepackage{times}
\usepackage{epsfig}
\usepackage{graphicx}
\usepackage{amsmath}
\usepackage{amssymb}
\usepackage[table]{xcolor}
\usepackage{bbm}
\usepackage{booktabs}
\usepackage[T1]{fontenc}
\usepackage{fix-cm}
\usepackage{array}
\iccvfinalcopy
\usepackage[pagebackref=true,breaklinks=true,letterpaper=true,colorlinks,bookmarks=false]{hyperref}

\def\iccvPaperID{576} 

\newcommand{\ve}[1]{{\mathbf #1}} 
\newcommand{\hua}[1]{{\mathcal #1}}

\graphicspath{{./fig/}}

\ificcvfinal\fi
\begin{document}

\title{Joint Object and Part Segmentation using Deep Learned Potentials}

\author{
Peng Wang$^{1}$~~Xiaohui Shen$^{2}$~~Zhe Lin$^{2}$~~Scott Cohen$^{2}$~~Brian Price$^{2}$~~Alan Yuille$^{1}$\\
$^{1}$University of California, Los Angeles~~~~$^{2}$Adobe Research\\
}

\maketitle

\begin{abstract}
   Segmenting semantic objects from images and parsing them into their respective semantic parts are fundamental steps towards detailed object understanding in computer vision. In this paper, we propose a joint solution that tackles semantic object and part segmentation simultaneously, in which 
   higher object-level context is provided to guide part segmentation, and more detailed part-level localization is utilized to refine object segmentation. Specifically, we first introduce the concept of semantic compositional parts (SCP) in which similar semantic parts are grouped and shared among different objects. A two-channel fully convolutional network (FCN) is then trained to provide the SCP and object potentials at each pixel. At the same time, a compact set of segments can also be obtained from the SCP predictions of the network. Given the potentials and the generated segments, in order to explore long-range context, we finally construct an efficient fully connected conditional random field (FCRF) to jointly predict the final object and part labels. Extensive evaluation on three different datasets shows that our approach can mutually enhance the performance of object and part segmentation, and outperforms the current state-of-the-art by a large margin on both tasks.
\end{abstract}

\vspace{-1.5\baselineskip}
\section{Introduction}
\vspace{-0.5\baselineskip}
\label{sec:intro}
Decomposing an object into semantic parts enables a more detailed understanding of the object, which can provide additional information to benefit many computer vision tasks such as pose estimation~\cite{DBLP:conf/cvpr/YangR11, DBLP:conf/cvpr/DongCSYY14}, detection~\cite{DBLP:conf/eccv/AzizpourL12, DBLP:conf/cvpr/ChenMLFUY14}, segmentation~\cite{DBLP:conf/nips/EslamiW12}, and fine-grained recognition~\cite{DBLP:conf/eccv/ZhangDGD14}. Thus, it has become an attractive research topic to leverage semantic part representation through part detection~\cite{DBLP:conf/iccv/BourdevM09, DBLP:journals/pami/FelzenszwalbGMR10, DBLP:conf/cvpr/ChenMLFUY14}, and human joint estimation~\cite{DBLP:conf/cvpr/ToshevS14}.

In the literature of semantic segmentation, while object-level segmentation over multiple object categories has been extensively studied along with the growing popularity of standard evaluation benchmarks such as PASCAL VOC~\cite{Database_VOC}, object parsing (i.e., segmenting objects into semantic parts) is addressed mostly for a few specific categories provided with accurate localization such as human~\cite{DBLP:journals/ijcv/ZhuCLY11, DBLP:conf/nips/EslamiW12, DBLP:conf/cvpr/DongCSYY14} and cars~\cite{DBLP:conf/nips/EslamiW12}.

\begin{figure}
\begin{center}
\includegraphics[width=\linewidth]{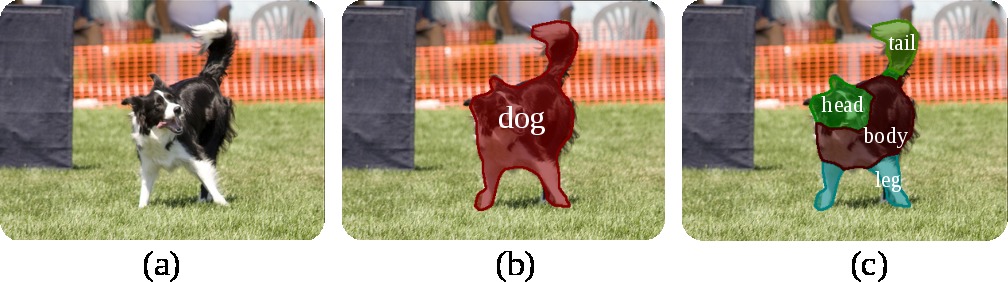}
\end{center}
\vspace{-1\baselineskip}
   \caption{We handle the prediction of semantic object and part segmentation in a wild scene scenario. (a) Original image. (b)$\&$(c) are the object and part segmentation respectively, generated from our algorithm.}
   \vspace{-1.8\baselineskip}
\label{fig:task}
\end{figure}

With the increasing availability of semantic part  annotations~\cite{DBLP:conf/cvpr/ChenMLFUY14}, more recent works have attempted to handle more difficult classes like animals with homogeneous appearance~\cite{Jianyu_CVPR15}, and to perform both object and part segmentation~\cite{BharathCVPR2015}, as illustrated in Fig.~\ref{fig:task}. However, in~\cite{BharathCVPR2015}, object and part segmentation are performed sequentially, in which the object mask is first segmented, and then the part labels are assigned to the pixels within the mask. As a result, the errors from the predicted semantic object masks may be propagated to the parts. 

In fact, object and part segmentation are complementary and mutually beneficial to each other. Semantic object segmentation requires a larger receptive field in order to correctly recognize the object, while part segmentation focuses on local details to obtain more accurate segmentation boundaries and accommodate large pose and viewpoint variations. If these two tasks are tackled simultaneously, by integrating the object-level guidance with part-level detailed segmentation, we can address two of the most challenging problems at the same time, i.e., discovering the subtle appearance differences between different parts within a single object, and avoiding the ambiguity across similar object categories. Motivated by this observation, we propose a joint solution to object and part segmentation, in which the consistency of the object and parts are enforced through joint training and inference.


\begin{figure*}[t]
\begin{center}
\includegraphics[width=\linewidth]{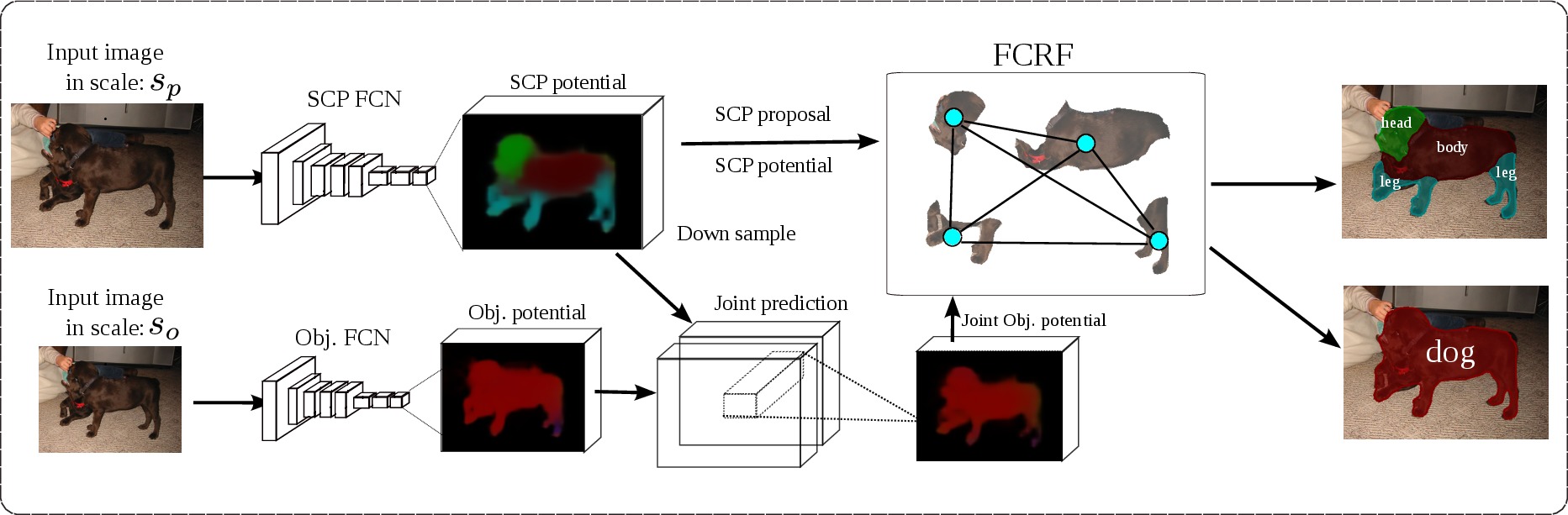}
\end{center}
   \caption{Our framework for joint part and object segmentation. Given an image, a two-channel FCN is performed to predict both the semantic compositional parts (SCP) and object potentials with different input image scales. The two potentials are then concatenated and fed to a new convolutional layer to predict joint object potentials. Finally, from SCP potentials, SCP segments are proposed as nodes for the fully-connected CRF to jointly infer the part and object labels.}
\label{fig:framework}
 \vspace{-1.3\baselineskip}
\end{figure*}
Fig.~\ref{fig:framework} shows the overall framework of our approach. When performing part segmentation over multiple object classes, the appearance of some parts may be very similar, \emph{e.g.}, horse legs and cow legs. Therefore in order to reduce the ambiguity and complexity during training, instead of treating each semantic part type independently~\cite{BharathCVPR2015}, we allow some part labels to be shared by related object classes, and group the labelled parts of different classes into a semantic compositional part (SCP) representation based on their appearance and shape similarity (\emph{e.g.}, horse legs and cow legs belonging to the same type of leg). Details of part sharing and SCP labels are illustrated in Fig.~\ref{fig:partsharing}. Since we are performing object and part segmentation jointly, the ambiguity of shared part labels can be solved by the object label.

Given an image with certain objects, we train a two-channel fully convolutional network (FCN) with the first channel predicting the SCP potentials while the second predicting the object potentials. We then concatenate the two potentials as the input to an additional convolutional layer to refine the object potentials through joint training. At the same time, a compact set of SCP region proposals  are generated from the SCP potentials. Using these region proposals as nodes, we construct a fully connected conditional random field (FCRF) to further incorporate the object and SCP potentials, yielding the jointly predicted object and part segmentations. In our FCRF, the consistency between the object and parts are enforced with long-range constraints. 

We did extensive experiments over three derived datasets based on PASCAL VOC segmentation benchmark~\cite{Database_VOC2010, DBLP:conf/cvpr/ChenMLFUY14}. Experimental results demonstrate that our joint approach improves both object and part segmentation, and significantly outperforms the state-of-the-art on both tasks. 

\vspace{-0.5\baselineskip}
\section{Related work}
\vspace{-0.5\baselineskip}
	In the literature of detection, the usefulness of mining semantic part representation in helping object recognition has been long studied. Felzenszwalb et.al~\cite{DBLP:journals/pami/FelzenszwalbGMR10, DBLP:conf/nips/GirshickFM11} proposed the deformable part-based model (DPM) which is an implicit way of discovering hidden parts.
Later models use more accurate part-representation through explicit part supervision~\cite{DBLP:conf/iccv/SunS11,DBLP:conf/cvpr/ZhuCTFY10}. Poselets~\cite{DBLP:conf/iccv/BourdevM09} are proposed to model the local human parts through 2D projections from 3D data, which can be used as robust representation for both detection~\cite{DBLP:conf/eccv/GuALYM12} and segmentation~\cite{DBLP:conf/iccv/MaireYP11, DBLP:conf/cvpr/ArbelaezHGGBM12}. In addition to human body parts, Azizpour et. al~\cite{DBLP:conf/eccv/AzizpourL12} explicitly induce  the bounding-box annotations of animal parts, yielding stronger detection results on both object and parts. Chen et. al~\cite{DBLP:conf/cvpr/ChenMLFUY14} extend such ideas by providing richer labeling of part segments, which can better capture the appearance features for learning.

In the literature of segmentation, semantic parsing has also been actively investigated. However, due to its increased challenge in getting detailed boundaries, most previous work focused on parsing objects given both the semantic category and a cropped bounding box with no occlusion, such as human parsing~\cite{DBLP:conf/cvpr/BoF11, DBLP:journals/ijcv/ZhuCLY11, DBLP:conf/cvpr/YamaguchiKOB12, DBLP:conf/nips/EslamiW12, DBLP:conf/cvpr/DongCSYY14}, car parsing~\cite{DBLP:conf/rss/ThomasFLTG08, DBLP:conf/nips/EslamiW12, DBLP:conf/bmvc/LuLY14} or animal parsing~\cite{Jianyu_CVPR15}. 
Such methods are limited in their applications, as objects in real-world images are often occluded with large deformation and appearance variations, which is difficult to be handled by those shape-based~\cite{DBLP:conf/cvpr/BoF11} or appearance-based~\cite{DBLP:conf/cvpr/DongCSYY14} models with hand-crafted features or bottom-up segments. 

Recently, deep convolutional neural networks~\cite{DBLP:conf/nips/KrizhevskySH12} (CNN) have achieved great success in many applications such as object detection~\cite{Girshick_2014_CVPR, DBLP:conf/eccv/HariharanAGM14, Yukun_CVPR15} and end-to-end segmentation~\cite{WangCVPR15, Eigenarxiv15, BharathCVPR2015, long_shelhamer_fcn, ChenICLR15}, with advanced network structures such as the VGG-Net~\cite{Simonyan14c}. Some studies tried to understand the implicitly learned filters~\cite{DBLP:conf/eccv/ZeilerF14} or comparable structures with the DPM~\cite{Wanarxiv14, girshick2015dpdpm} in the network, and discovered some meaningful part representations in deeper layers. However, such representations are still not semantic parts, and using explicit supervision from semantic part labels for segmentation has not been investigated. 
In our approach, we propose to explicitly model the semantic parts along with the whole object by taking advantage of the recent advance of fully convolutional network (FCN)~\cite{long_shelhamer_fcn}. It has been very successful in predicting structured output such as semantic object segmentation. Specifically, FCN converts the fully-connected layers in the original CNN to $1\times 1$ convolution layers, thus can efficiently perform sliding-window-based classification at each pixel with a certain receptive field. However, it starts from local convolutional kernels with limited receptive fields and may not be able to capture all the long-range context, yielding local confusions. In our case, we solve such problem through modelling over object scale context, which is also suggested in prior arts~\cite{DBLP:conf/iccv/BourdevM09, DBLP:conf/cvpr/DivvalaHHEH09, DBLP:journals/pami/TuB10, DBLP:conf/eccv/DoerschGE14}. 

Perhaps the closest work in our scenario is the hypercolumn approach~\cite{BharathCVPR2015}. However, they perform object and part segmentation sequentially, and train many part classifiers separately for each class, which may suffer from increased training cost and has less scalability when the number of object class is large. In contrast, we use semantic compositional part (SCP) to allow part sharing and reduce training complexity. Moreover, our model leverages the advantage of both object segmentation and semantic part segmentation, yielding strong results in very challenging scenarios.
To the best of our knowledge, this is the first work that provides a joint solution to tackle the segmentation of semantic parts and objects, which allows the part and object potentials interact and benefit each other.
	
	
	
\vspace{-0.5\baselineskip}	
\section{Joint part and object segmentation}
\vspace{-0.5\baselineskip}

Our framework includes four major parts, i.e. shared semantic compositional parts (SCP) generation,  part and object potentials,  proposal of SCP regions and fully connected conditional random field (FCRF). In the following sessions, we will describe these techniques in details. 

\vspace{-0.3\baselineskip}
\subsection{Semantic compositional parts}
\vspace{-0.2\baselineskip}
\label{subsec:grammar}
When performing semantic object segmentation and parsing over multiple object classes, some parts from different object classes yet with similar semantic meanings may also have very similar shapes and appearances (\emph{e.g.}, horse legs and cow legs). In such scenarios, allowing the parts to be shared among similar object classes~\cite{DBLP:conf/cvpr/ZhuCTFY10,DBLP:conf/cvpr/OttE11} could alleviate the difficulties of distinguishing similar parts from different objects, and at the same time reduce the increasing complexity of training and inference as the number of object categories grows. Therefore, before we formally train our framework for these two tasks, we group those similar parts to form the semantic compositional parts (SCP) that are shared among related object classes. 

\begin{figure}[t]
\vspace{-0.3\baselineskip}
\begin{center}
\includegraphics[width=\linewidth]{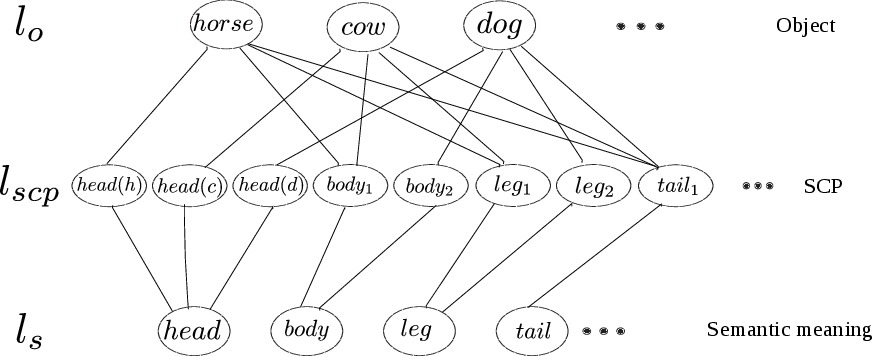}
\end{center}
   \caption{Illustration of our semantic compositional part (SCP) grammer in Sec.~\ref{subsec:grammar}. Each SCP is associated with one semantic meaning, and all the objects are composed of several SCPs.}
\vspace{-1.3\baselineskip}
\label{fig:partsharing}
\end{figure}

In particular, given a semantic part, it has an object label $l_o$ (\emph{e.g}., horse) and a particular semantic meaning $l_s$ (\emph{e.g}, leg). The joint label of this part we would like to infer is denoted by $l_{op}$ (\emph{e.g}, horse-leg). We group the original part labels $l_{op}$ to a shared compositional part representation $l_{scp}$  if they have the same semantic meanings $l_s$ and highly similar appearances and shapes.  The SCP $l_{scp}$ are then used to compose different objects $l_o$ as illustrated in Fig.~\ref{fig:partsharing}. For example, in the case of horse and cow, the representation of different labels are, $l_o \in \{$horse, cow$\}$,  $l_{op}\in\{$horse-head, horse-body, horse-leg, horse-tail, cow-head, cow-leg, cow-body, cow-tail$\}$ and $l_s\in\{$head, body, leg, tail$\}$. If we allow the two objects to share the same type of body, leg and tail, we get the SCP label $l_{scp}\in\{$head(h), head(c), body$_{1}$, leg$_{1}$, tail$_{1}$ $\}$, as in Fig.~\ref{fig:partsharing}, which is a much smaller prediction space than $l_{op}$. The information of $l_{op}$ is kept in the connections between $l_o$ and $l_{scp}$. During the inference, by enforcing the consistency of
$l_{scp}$ and $l_o$, $l_{op}$ can be directly recovered, \emph{e.g}. known $l_o=$ horse and $l_{scp}=$ leg$_1$, then we get $l_{op}=$ horse-leg. 
Currently we manually group those part labels. Nonetheless, automatically generating SCP is a very interesting problem especially with increased number of object categories, and will be investigated in the future.

\vspace{-0.3\baselineskip}
\subsection{Deep part and object potentials}
\label{subsec:potential}
\vspace{-0.3\baselineskip}

In this section, we mainly describe the joint prediction of the semantic compositional parts (SCP) potentials and object potentials by a two-channel FCN, which is then used to construct the FCRF. Specifically, as illustrated in the framework (Fig.~\ref{fig:framework}), the first channel of the FCN predicts a $(N_{p}+1)$-channel SCP potential map ($N_{p}$ is the SCP label number, while the additional label represents the background). Similarly, the second channel of the network predicts a $(N_o+1)$-channel object-class potential map ($N_{o}$ being the object class number). In addition, we concatenate the SCP potentials and the object potentials as a set of high-level features, and feed them to a new convolutional layer for object potential refinement. This predicted joint object potential has less noise within the object and better boundaries, as the SCP potentials contain more fine-level details that can interact with the object potentials during joint training. As shown in our experiments (Sec.~\ref{subsec:benchmark}), compared with using the original object potentials from FCN, our joint object potentials provide better evidence for the graphical model later, yielding better final results. 

One may consider to also generate refined SCP potentials similarly using the joint prediction layer. However, using the SCP potentials in this way does not show much improvement in our experiments. This is because firstly, the SCP potentials have already encoded more detailed boundary information than the object potentials, and the interactions would not help the SCP potentials to refine their boundaries. Secondly, the ambiguity of similar parts from different objects, which is the most challenging problem in part segmentation, has already been better addressed by using part sharing and the object-scale FCRF. Therefore, the joint prediction for SCP refinement is not adopted in our framework to reduce the system complexity. 



Last but not the least, the SCP potentials and object potentials actually need different levels of context. This is also a key factor in DPM~\cite{DBLP:journals/pami/FelzenszwalbGMR10}, where they use a larger image scale for part filters and a smaller scale for root filters. In our case, we adopt a similar strategy with different input resolutions to obtain proper receptive fields, i.e. $s_p\times s_p$ for SCP and $s_o\times s_o$ for object, where we require $s_p > s_o$. We investigated the influence of input scale in our experiments, and chose the optimal $s_p$ and $s_o$ using cross-validation. 


\vspace{-0.3\baselineskip}
\subsection{SCP segments proposal}
\label{subsec:SCP}
\vspace{-0.3\baselineskip}

As mentioned earlier, our FCRF is built upon a compact set of SCP segments. In our case, traditional object proposal algorithms such as  CPMC~\cite{DBLP:dblp_conf/eccv/CarreiraCBS12} or MCG~\cite{DBLP:conf/cvpr/ArbelaezPBMM14} would typically fail due to the subtle difference of appearance between connected parts such as the leg and body. Nevertheless, we can use the SCP FCN network to generate accurate segments that are associated with SCP. Based on the predicted $(N_{p}+1)$-channel SCP probability map, we assign the SCP label with the highest probability to each pixel and generate the SCP label map. SCP segments are then generated by grouping the pixels with the same SCP labels.

\begin{figure}[t]
\begin{center}
\includegraphics[width=\linewidth]{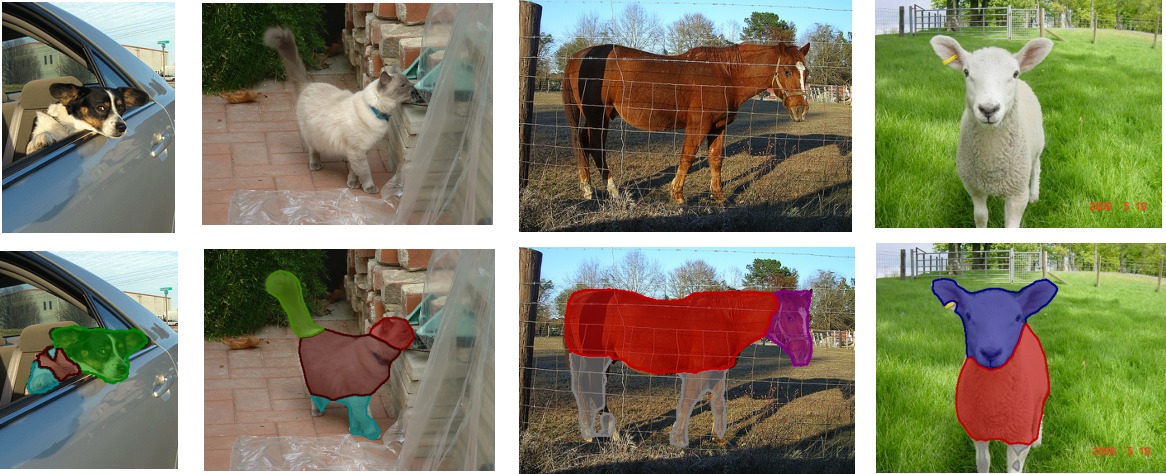}\\
\vspace{5pt}
\includegraphics[width=\linewidth]{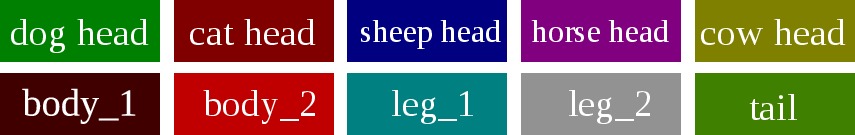}
\end{center}
   \caption{The SCP segments proposals handle various difficulties such as truncation, occlusion, deformation and view point changes. The colors maps shown  at the bottom for each kind of SCP. We keep the color consistent in the results.}
\vspace{-1.2\baselineskip}
\label{fig:scp}
\end{figure}

Fig.~\ref{fig:scp} shows several examples of the proposed SCP segments from an image. We can see our SCP segments work very well in terms of capturing correct semantic part regions and locating the object boundaries. To practically evaluate the segments, we did an oracle experiments by assigning the proposed SCP segments to be the overlapping ground truth class labels. The best possible object IOU is $85.1\%$ over the quadrupeds animal set in Sec.~\ref{subsec:benchmark}, which performs reasonably well for capturing the object boundaries. One might also apply local dense pixel-wise refine strategy~\cite{ChenICLR15} to further refine the segments, but it will increase computational cost and need further investigation.

\vspace{-0.3\baselineskip}
\subsection{Joint FCRF}
\label{subsec:FCRF}
\vspace{-0.3\baselineskip}
FCN is essentially a sliding-window based approach, where its receptive field is always fixed to a local area given the input image, thus local confusion can hardly be avoided in many cases.
Intuitively, the optimal receptive field should be close to the scale of the presented object in the image. Therefore, after generating SCP segments as well as SCP and object potentials from the joint FCN, we further construct a fully-connected CRF (FCRF) to automatically discover the object-scale context, where 
all the parts of an object can interact with each other, yielding the optimal semantic object and part prediction simultaneously.

Specifically, given the set of SCP segment proposals, we first cluster the proposed SCP segments into several groups according to their spatial distances, in case there are multiple isolated objects in an image. Two SCP segments are merged in the same group if the minimum distance of the pixels within these two segments is smaller than a threshold $t_s = 10$. We assume each group of SCP segments forms an object or overlapping objects in the image, which provides a good estimation of the object-scale. Then, we build one FCRF for each corresponding group. Formally, the FCRF can be represented as $\hua{G} = \{\hua{V}, \hua{E}\}$, where $\hua{V}$ is the set of SCP segments in the same group and $\hua{E}$ is the set of edges connecting every pair of segments. 
As introduced in our framework (Fig.~\ref{fig:framework}), for each SCP segment $\ve{P}$, we want to predict its semantic part label $l_{op}(\ve{P})$ (defined in Sec.~\ref{subsec:grammar}), but can be reduced to separately inferring the object label $l_o(\ve{P})$ and SCP label $l_{scp}(\ve{P})$ by enforcing their label consistency. 
In the following we use $l_{op}^i$ for $l_{op}(\ve{P}_i)$, $l_{p}^i$ for $l_{scp}(\ve{P}_i)$ and $l_o^i$ for $l_o(\ve{P}_i)$ for simplicity.

In sum, our FCRF is formulated as, 
\begin{align}
\vspace{-1.5\baselineskip}
\label{eqn:energy}
&\min_{\hua{L}}\sum\limits_{i\in\hua{V}}\psi_{i}(l_{op}^i) + \lambda_{e}\sum\limits_{i, j \in \hua{V}, i\neq j} \psi_{i, j}(l_{op}^i, l_{op}^j) \nonumber \\
\mbox{where,~~} 
&\psi_{i}(l_{op}^i) = \eta(l_o^i, l_p^i)(\psi_{i}^o(l_o^i)) + \lambda_{p}\psi_{i}^p(l_p^i));  \\
&\psi_{i, j}(l_{op}^i, l_{op}^j) = \eta(l_o^i, l_p^i)\eta(l_o^j, l_p^j) \psi_{i, j}^{op}(l_o^i, l_o^j, l_p^i, l_p^j); \nonumber
\vspace{-0.5\baselineskip}
\end{align}
where $\hua{L}$ is the label set of the proposed SCP segments, $\lambda_{e}$ and $\lambda_{p}$ are balancing parameters. 
$\psi_{i}^o(l_o^i) = \sum_{\ve{x}_j\in \ve{P}_i}-\log(P(l_o(\ve{x}_j))$, is the sum of pixel-wise object potentials inside the SCP segment $\ve{P}_i$,  and $\psi_{i}^p(l_p^i)$ is similarly defined with the SCP potentials. $\eta(l_o^i, l_p^i)$ is a constraint for part and object combination, which is $1$ if $l_o^i, l_p^i$ is a meaningful combination, i.e. a connection exists between $l_o^i$ and $l_p^i$ in the grammar as in Fig.~\ref{fig:partsharing}, and set to be $\infty$ otherwise. 


In order to learn the pairwise potentials, we train a two-layer fully-connected neural network, which takes the features from a pair of segments $i$ and $j$ as input, and predicts the probabilities of the four labels $l_o^i$, $l_o^j$, $l_p^i$, $l_p^j$. The ground truth labels for each training segment are the most dominant labels of the pixels within the segment, and we adopt the multinomial logistic loss for training.

For the pairwise features, we consider multiple semantically meaningful cues from the pairwise relations, i.e. $\ve{f}_{ij} = [\ve{f}_{i}^T, \ve{f}_{j}^T,  \kappa_{ij}, \ve{\theta}_{j|i}]^T$, where $\ve{f}_{i} = [\ve{f}_{oi}^T, \ve{f}_{pi}^T, \ve{f}_{ai}]^T$ is a segment self-descriptor, $\kappa_{ij} = [d^a_{ij}, d^p_{ij}]^T$ is a spatial metric of the segment pair. $\ve{\theta}_{j|i}$ is the relative angle. 
We summarize the features as follows, 
\begin{itemize}
\vspace{-0.4\baselineskip}
\setlength\itemsep{1pt}
\addtolength{\itemindent}{-0.2cm}
\item $\ve{f}_{oi}^T$: the mean of pixel-wise object potentials. 
\item $\ve{f}_{pi}^T$: the mean of pixel-wise SCP potentials.
\item $\ve{f}_{ai}$: the segment area, normalized by the discovered object area.
\item $d^a_{ij}$: the appearance geodesic distance from segment $i$ to $j$, which accumulates the edge weights along the path from $i$ to $j$ on the image. 
\item $d^p_{ij}$: the Euclidean distance between the center of the two segments normalized by the height and width of the object. 
\item $\ve{\theta}_{j|i}$: the relative angle of the center of the segments $j$ with respect to the center of segment $i$.
\vspace{-0.4\baselineskip}
\end{itemize}
Specifically for $d^{a}_{ij}$, the edge weight between the two neighbouring segments is the sum of the edgemap~\cite{DBLP:dblp_conf/eccv/LeordeanuSS12} values over their overlapping boundary.

After the model is learned, given the pairwise feature $\ve{f}_{ij}$, the potentials $\psi_{i, j}^{op}(l_o^i, l_o^j, l_p^i, l_p^j)$ is computed as the negative log-likelihood, i.e. $-\log(P(l_o^i)P(l_o^j)P(l_p^j)P(l_p^j))$, where the probabilities are from the neural network prediction.



For inference, since our graphical model have a very small number of nodes (less than 15 SCP segments in average), using the efficient LBP~\cite{DBLP:dblp_journals/corr/abs-1301-6725}, our algorithm can converge very fast within 5 iterations. 

\vspace{-0.3\baselineskip}
\subsection{Relation to the DPM structure}
\vspace{-0.3\baselineskip}
While we are dealing with semantic segmentation and parsing, our approach follows the spirit of DPM~\cite{DBLP:journals/pami/FelzenszwalbGMR10} in object detection. In~\cite{girshick2015dpdpm}, Girshick et. al connected the DPM root filter and part filters with the convolutional filters from CNN, and the distance transform can be regarded as an additional pooling and geometry filtering step.
Our work aims to solve the segmentation task here but has analogy with DPM, where the object prediction can be considered as the root filter and SCP potentials are our learned part filters. The difference is that rather than firing when the target is at the exact center of the detection window, our model fires at every location inside the target which produces more accurate location estimation. Regarding the part and root geometry, rather than explicitly modeling the geometry of root-part distance transform as in DPM, we implicitly model these spatial relationships using spatial distances and relative angles as spatial features and learn the pair-wise potentials in the fully-connected graphic model, which is more data-driven and generalizes better for handling variations.

For inference, DPM tries to find the probability of the part-root locations $\ve{p}$ given the object label $l_o$ through sliding window, i.e. $\max_{\ve{p}\in\ve{I}}P(\ve{p} | l_o)$. In our case, sliding-window for part and object localization is realized by our FCN, based on which we can infer over the smaller label space i.e. $\max_{l_o\in\hua{L}}P(l_o|\ve{p})$, at the object-scale context.

\vspace{-0.5\baselineskip}
\section{Experiments}
\vspace{-0.5\baselineskip}
\label{sec:exp}
In this section, we provide all the experimental details, and evaluate our approach in terms of different experimental settings to demonstrate the advantage
of our approach. 
Specifically, we conduct experiments on the Horse-Cow parsing dataset introduced in~\cite{Jianyu_CVPR15}, the PASCAL Quadrupeds dataset from~\cite{DBLP:conf/cvpr/ChenMLFUY14}, and our PASCAL Part benchmark, over which extensive comparisons are performed. 
\vspace{-0.5\baselineskip}
\subsection{Model training.}
\vspace{-0.5\baselineskip}
\label{sec:training}
Both of our FCN models for SCP and object prediction are based on the 16-stride (16s) FCN, since there is trivial improvement (less than $1\%$ as shown in our Tab.~\ref{tab:resAnimal} and Tab.2 of~\cite{long_shelhamer_fcn}), while significant more time required to train a 8-stride model (8s).
\vspace{-1\baselineskip}
\paragraph{Data augmentation.}
Effective data augmentation is the key to the success of FCN. Thus in our case, we did sufficient augmentation by first cropping out the connected object masks using a random generated bounding box around it. The size of the crop is specially 1.3 times larger than the object bounding box, which is a rough localization of the object for generalization. Then, each cropped image is resized into $300\times 300$, based on which we further augmented using the ideas from~\cite{Eigenarxiv15}, i.e. we perform 4 additional random cropping at the size of $200\times 200$, flipping, changing the color intensity by a random scale in $[0.7,1.3]$ with a  probability of $0.4$ and rotation in $[-5,+5]$ degree with a probability of $0.5$. In average, each image is augmented up to around 25 training samples. 
\vspace{-\baselineskip}
\paragraph{Optimization and step-wise training.} 
We fine-tune all our models step-by-step from the publicly available VGG-net~\cite{long_shelhamer_fcn}. For training the SCP FCN, we first train a 32s FCN with  the learning rate as $10^{-4}$ for the final $conv_{fc8}$ layer which is the layer predicting the SCP potentials, and $10^{-5}$ learning rate for the layers after $pool4$, while we fix the layers before $pool4$. Then for the 16s FCN, we start with training the $conv_{poo4}$ layer which is used for predicting SCP potentials from the $pool4$'s output. Then, we use the trained layers as initialization for the final 16s FCN training. We further fine-tune the $conv_{fc8}$ and $conv_{poo4}$ layer with a learning rate as $10^-5$. For training the joint object FCN, we fix the SCP FCN and concatenate the SCP potentials with the 16s object potentials, over which another convolutional layer $conv_{jnt}$ (with a kernel size of $5\time 5$) is used to predict the joint object potentials. For fine-tuning this model, we use $10^{-4}$ learning rate for the $conv_{jnt}$ layer, while $10^{-6}$ learning rate for the $conv_{fc8}$ and $conv_{pool4}$ of the object FCN. In all the cases, we keep the batch size as 32. For the other parameters, we refer to the ones given by~\cite{long_shelhamer_fcn}.

For the two-layer neural network in training the pairwise potentials of the fully-connected CRF, we set $32$ hidden nodes, and use the RELU for no-linearity with a dropout rate of $0.2$ to regularize the model. We use a batch size of $10000$ and set the learning rate to be $10^{-2}$. In learning the pair-wise term, other than separately training SCP and object labels, i.e. $l_{scp}$ and $l_{o}$, one may also consider output the semantic part label $l_{op}$ which lies in the joint space of object and part. Nevertheless, we found it is harder to train due to that a lot more data are required for prediction in a high dimensional output space. Thus we chose to train separately. 

All our neural networks are based on the caffe platform~\cite{jia2014caffe} and partially from the code provided by~\cite{long_shelhamer_fcn}. 
\vspace{-0.3\baselineskip}
\subsection{Parameters and details}
\begin{figure}[t]
 \vspace{-0.5\baselineskip}
\begin{center}
\includegraphics[width=1\linewidth]{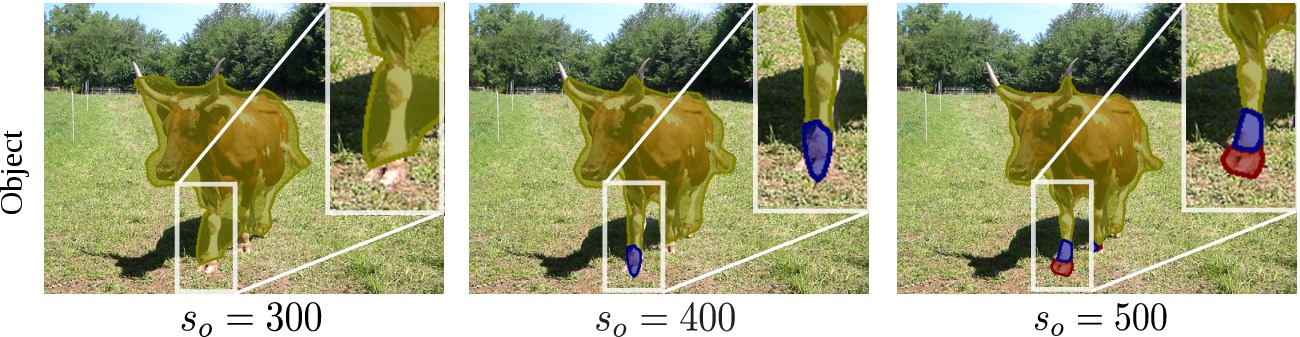}\\
\vspace{5pt}
\setlength{\tabcolsep}{10pt}
\begin{tabular}{|l|c|c|c|}
\hline
&$s_p=400$&$s_p=500$&$s_p=600$\\
\hline
$s_o=300$&78.58&78.88&\textbf{79.18}\\
\hline
$s_o=400$&77.55&78.62&78.11\\
\hline
$s_o=500$&77.09&75.81&74.20\\
\hline
\end{tabular}
\end{center}
   \caption{Investigation of image scales. Top: an example from our joint object FCN prediction, showing that larger scale leads to finer boundaries, but introduces local ambiguities. Bottom: validate the accuracy of scale combination of $s_o$ and $s_p$.}
\label{fig:scale}
\vspace{-1.3\baselineskip}
\end{figure}

\vspace{-0.5\baselineskip}
In the FCRF, we set $\lambda_e = 2$ and $\lambda_p = 0.3$ which are validated over a validation set from the Quadrupeds dataset. The same set is used for all other validation experiments. For inference over the graphical model, we use the LBP tool provided by Meltzer\footnote{http://www.cs.huji.ac.il/~talyam/inference.html}.
\vspace{-0.8\baselineskip}
\paragraph{Investigation of input image scale.}
The input image scale for FCN is one of the most important factors for achieving good performance as also shown in recent works~\cite{ChenICLR15,LINarxiv15}. Fig.~\ref{fig:scale} shows the investigation of changing the input image scale for SCP proposals and object potentials. In the top row, we can see the larger the image scale is, the more accurate the object boundaries could be, while the more local confusion it has, \emph{e.g.} a leg of the cow is starting to be confused with horse. Thus, we validate the scale of $s_p\in \{400, 500, 600\}$ and $s_o \in \{300, 400, 500\}$. At the bottom of Fig.~\ref{fig:scale}, we show the validated results, and the optimal combination, i.e. $s_p = 600, s_o = 300$ is used in all our experiments. 


\vspace{-0.3\baselineskip}
\paragraph{Training and inference time.} For training, we found the model would converge after 40k iterations and it takes around 2 days for SCP FCN, 1 day for object potentials and 8 hours for the graphical model with a platform of 4 core 3.2Hz CPU and a K40 GPU. For inference, in average, one image takes 0.3s for FCN forward propagation with the GPU and 1.3s for the graph model with our CPU. 
\vspace{-0.3\baselineskip}
\subsection{Performance comparisons}
\vspace{-0.5\baselineskip}
\label{subsec:benchmark}
We compare our algorithm with three state-of-the-art methods on the two tasks. For semantic part segmentation, we compare with the most recent compositional-based semantic part segmentation (SPS)~\cite{Jianyu_CVPR15} over the Horse-Cow parsing dataset, and the hypercolumn (HC)~\cite{BharathCVPR2015} in all datasets. For object segmentation, we compare our method with the FCN~\cite{long_shelhamer_fcn}. We use the code provided by the author and fine-tune their model based on our dataset, including tuning a fixed optimal image scale for a fair comparison. For comparing with HC~\cite{BharathCVPR2015}, as there is no available code from the author, we follow their part parsing strategy by first performing figure-ground mask with the trained FCN 8s~\cite{long_shelhamer_fcn}, and then assign part labels inside with a optimally tuned image scale to be our baseline method. For evaluation, we adopt the standard intersection over union (IOU) criteria for both tasks.	
\begin{figure}[b]
\begin{minipage}[t]{0.9\linewidth}
\flushleft
   \vspace{-1.3\baselineskip}
\begin{tabular}{rc@{~}c@{~}c}
\rotatebox{90}{Image} &
\includegraphics[width=0.21\linewidth, height=0.29\linewidth]{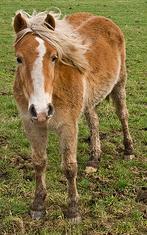} &
\includegraphics[width=0.35\linewidth, height=0.28\linewidth]{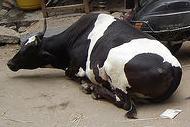} &
\includegraphics[width=0.35\linewidth, height=0.28\linewidth]{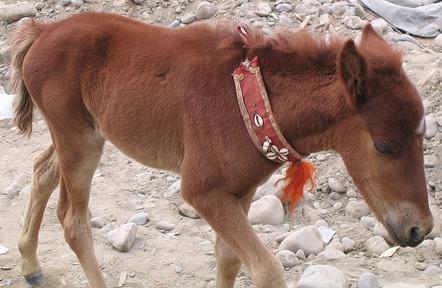} \\
\rotatebox{90}{SPS~\cite{Jianyu_CVPR15}}& 
\includegraphics[width=0.21\linewidth, height=0.29\linewidth]{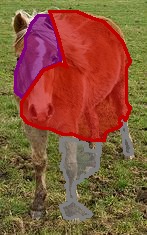} &
\includegraphics[width=0.35\linewidth, height=0.28\linewidth]{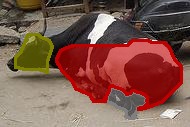} &
\includegraphics[width=0.35\linewidth, height=0.28\linewidth]{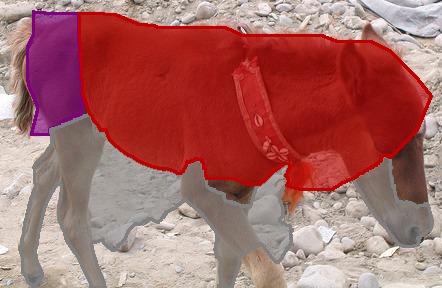} \\
\rotatebox{90}{\small{HC~\cite{BharathCVPR2015}}} & 
\includegraphics[width=0.21\linewidth, height=0.29\linewidth]{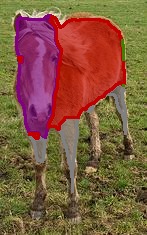} &
\includegraphics[width=0.35\linewidth, height=0.28\linewidth]{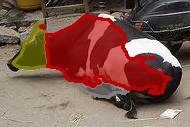} &
\includegraphics[width=0.35\linewidth, height=0.28\linewidth]{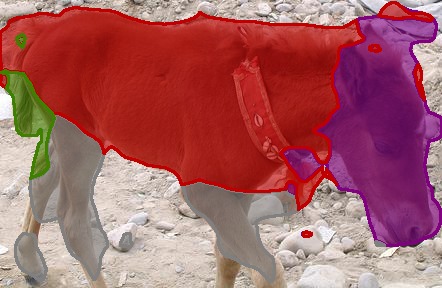} \\
\rotatebox{90}{Ours} & 
\includegraphics[width=0.21\linewidth, height=0.29\linewidth]{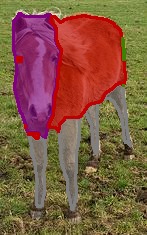} &
\includegraphics[width=0.35\linewidth, height=0.28\linewidth]{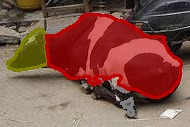} &
\includegraphics[width=0.35\linewidth, height=0.28\linewidth]{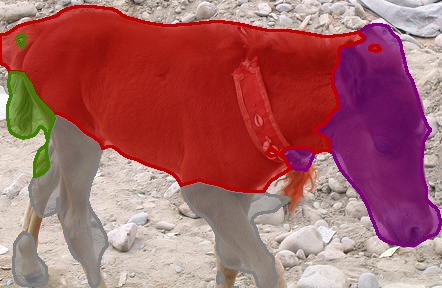} \\
\rotatebox{90}{Part GT.} &
\includegraphics[width=0.21\linewidth, height=0.29\linewidth]{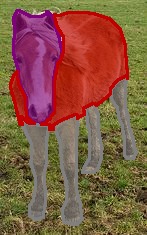} &
\includegraphics[width=0.35\linewidth, height=0.28\linewidth]{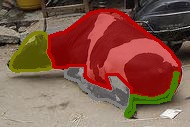} &
\includegraphics[width=0.35\linewidth, height=0.28\linewidth]{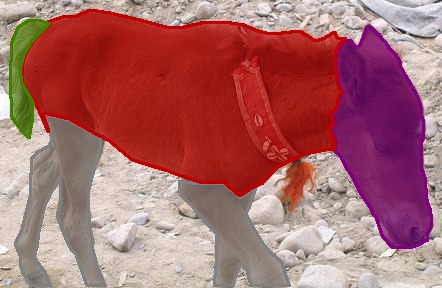} \\
\end{tabular}
\caption{Comparison examples with SPS~\cite{Jianyu_CVPR15} and HC~\cite{BharathCVPR2015} in the horse, cow dataset and the color map is in Fig.~\ref{fig:scp} (Best view in color).}
\label{fig:horsecow}
\end{minipage}
\end{figure}

\vspace{-1.2\baselineskip}
\paragraph{Horse-Cow parsing dataset.}
The Horse-Cow dataset is a two-animal part segmentation benchmark proposed in~\cite{Jianyu_CVPR15}. For each animal class, they manually select the mostly observable animal instances in both train-val and test set from the PASCAL VOC 2010~\cite{Database_VOC}.  There are 294 training examples and 277 test images. Since it is not published yet, we asked the author for this dataset and their results. In~\cite{Jianyu_CVPR15}, the task is to segment the part given the object class. Thus for a fair comparison, we also test every instance with the object class known. In such a case, our graphical model in Eqn.(\ref{eqn:energy}) is reduced to inferring the part label $l_p$ with a binary object potential $l_o$.

Tab.~\ref{tab:horseCow} provides the results of SPS~\cite{Jianyu_CVPR15}, HC~\cite{BharathCVPR2015} and our method over semantic part and figure-ground segmentation. Our method outperforms the previous state-of-the-art with a significant margin, averagely $13\%$ better than SPS and $4.5\%$ better than HC. 
Several qualitative examples are shown in Fig.~\ref{fig:horsecow}, where we keep the color map consistent with the SPC labels (Fig.~\ref{fig:scp}). In these cases, we can see that, built on explicit geometry rules, SPS is limited in handling large variance of object parts, which makes it difficult to model highly deformable cases like the tail, or occluded animals. For HC, due to inaccuracy from the object mask, it may miss some detailed regions like legs (the $1_{st}$ column of horse). We will further compare to HC in  later benchmarks. 

\setlength{\tabcolsep}{4pt}
\begin{table}[t]
\scriptsize
\centering
\rowcolors{2}{}{gray!35}
\begin{tabular}{ l c c c c c c c c }
\toprule[0.1 em] 
\toprule[0.1 em]
&\multicolumn{8}{c}{Horse}\\
& Bkg & head & body & leg & tail & Fg & IOU & Pix. Acc \\
\midrule 
SPS~\cite{Jianyu_CVPR15} & 79.14 & 47.64 & 69.74 & 38.85 & - & 68.63 & - & 81.45\\ 
HC~\cite{BharathCVPR2015} & 85.71 & 57.30 & \textbf{77.88} & 51.93 & 37.10 & 78.84 & 61.98 & 87.18\\
\midrule
Ours & \textbf{87.34} & \textbf{60.02} & 77.52 & \textbf{58.35} & \textbf{51.88} & \textbf{80.70} & \textbf{65.02}  & \textbf{88.49}\\
\bottomrule[0.1 em]
\end{tabular}
\rowcolors{2}{}{gray!35}
\begin{tabular}{ l c c c c c c c c }
\toprule[0.1 em]
&\multicolumn{8}{c}{Cow}\\
& Bkg & head & body & leg & tail & Fg & IOU  & Pix. Acc \\
\midrule
SPS~\cite{Jianyu_CVPR15} & 78.00 & 40.55 & 61.65 & 36.32 & - & 71.98 & - & 78.97\\
HC~\cite{BharathCVPR2015} & 81.86 & 55.18 & 72.75 & 42.03 & 11.04 & 77.07 & 52.57 & 84.43\\
\midrule
Ours & \textbf{85.68} & \textbf{58.04} & \textbf{76.04} & \textbf{51.12} & \textbf{15.00} & \textbf{82.63} & \textbf{57.18} & \textbf{87.00} \\
\bottomrule[0.1 em] 
\end{tabular}
 \vspace{0.3\baselineskip}
\caption{ Average precision over the Horse-Cow dataset.}
 \vspace{-2\baselineskip}
\label{tab:horseCow}
\end{table}

\vspace{-1\baselineskip}
\paragraph{Quadrupeds dataset.}
We further extend our experiments into the Quadrupeds part dataset which contains five animal classes, i.e., cat, dog, sheep, cow and horse. In this task, we simultaneously predict the object and part masks.
We obtain the data given by~\cite{DBLP:conf/cvpr/ChenMLFUY14}, which include all the part labels in PASCAL VOC 2010 training and validation images. We select the images containing the target objects, and treat the validation set as our test images, resulting in 3120 training images and 294 testing images.
In addition, since we are focusing more on semantic part segmentation, during testing, we roughly localize the object inside by using the strategy in data augmentation (Sec.~\ref{sec:training}). We will provide our code of this localization to others for a fair comparison. It should be noted that such localization is still very coarse with much looser bounding boxes than the ones needed in previous parsing methods~\cite{DBLP:conf/cvpr/DongCSYY14, Jianyu_CVPR15}. 

\setlength{\tabcolsep}{3pt}
\begin{table}[b]
\vspace{-1\baselineskip}
\scriptsize
\centering
\rowcolors{2}{}{gray!35} 
\begin{tabular}{l c c c c c c c c}
\toprule[0.1 em]
\toprule[0.1 em]
&\multicolumn{8}{c}{Object segmentation accuracy}\\
& Bkg & Dog & Cat & Cow & Horse & Sheep & IOU & Pix. Acc \\
\midrule
FCN 16s~\cite{long_shelhamer_fcn} & 93.25 &74.30 &78.62 &61.88 &56.56 &67.63 &72.04 & 93.00\\
FCN 8s~\cite{long_shelhamer_fcn} & 93.55 &74.39 &78.52 &60.81 &58.39 &69.15 &72.47 & 93.17\\
\midrule
Joint FCN(16s) & 94.04 &75.13 &80.52 &66.76 &63.04 &71.54 &75.17 & 93.77 \\
FCRF+FCN(16s) & 93.88 &77.10 &80.92 &68.76 &63.40 &64.54 &74.57 & 93.87 \\
Ours final & \textbf{94.40} & \textbf{79.03} & \textbf{83.04} & \textbf{74.82} & \textbf{69.94} & \textbf{70.59} & \textbf{78.64 } & \textbf{94.71}\\
\bottomrule[0.1 em]
\end{tabular}
\setlength{\tabcolsep}{4pt}
\rowcolors{2}{}{gray!35}
\begin{tabular}{ l c c c c c c c c }
\toprule[0.1 em]
&\multicolumn{8}{c}{Semantic part segmentation accuracy}\\
& Bkg & Dog & Cat & Cow & Horse & Sheep & IOU & Pix. Acc \\
\midrule
HC~\cite{BharathCVPR2015}   & 92.83 & 42.07 &43.99 &35.49 &38.59 &33.80 &41.36 & 89.54\\
\midrule
Ours final &  \textbf{94.46} & \textbf{45.63} & \textbf{47.81} & \textbf{42.7} & \textbf{49.60}& \textbf{35.74}& \textbf{46.69}& \textbf{91.74}\\
\bottomrule[0.1 em]
\end{tabular}
\caption{ Average precision over the Quadrupeds data.}
\vspace{-0.8\baselineskip}
\label{tab:resAnimal}
\end{table}

\begin{figure*}[t]
\vspace{-0.8\baselineskip}
\begin{center}
\begin{tabular}{c@{~}c@{~}c@{~}c@{~}c@{~}c@{~}c}
\multicolumn{4}{c}{\includegraphics[width=0.56\linewidth]{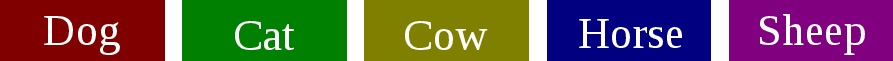}}&&&\\
\includegraphics[width=0.14\linewidth]{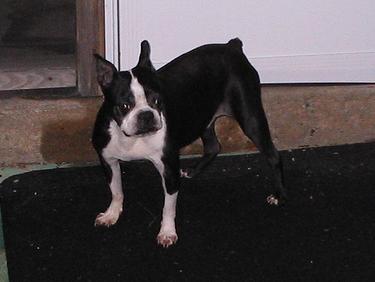}&
\includegraphics[width=0.14\linewidth]{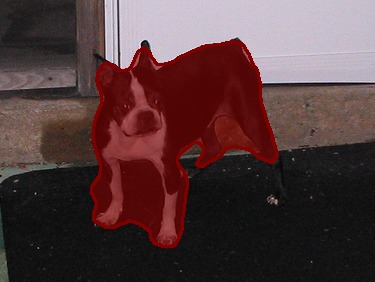}&
\includegraphics[width=0.14\linewidth]{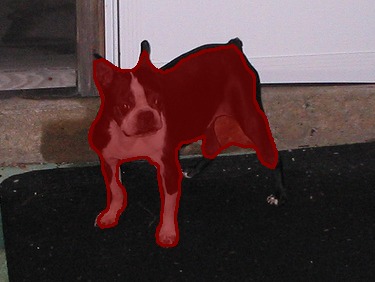}&
\includegraphics[width=0.14\linewidth]{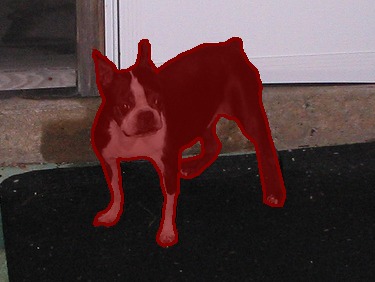}&
\includegraphics[width=0.14\linewidth]{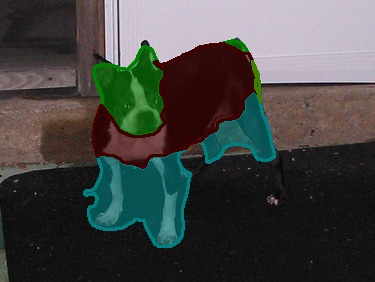}&
\includegraphics[width=0.14\linewidth]{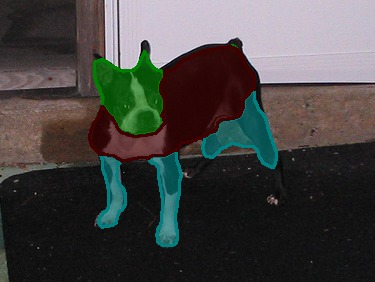}&
\includegraphics[width=0.14\linewidth]{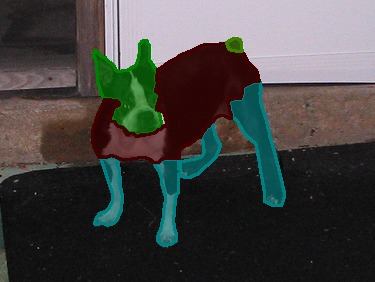}\\
\includegraphics[width=0.14\linewidth]{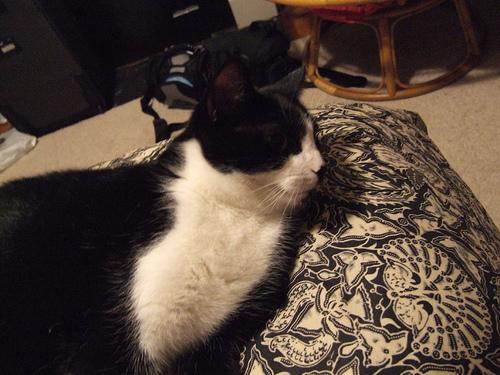}&
\includegraphics[width=0.14\linewidth]{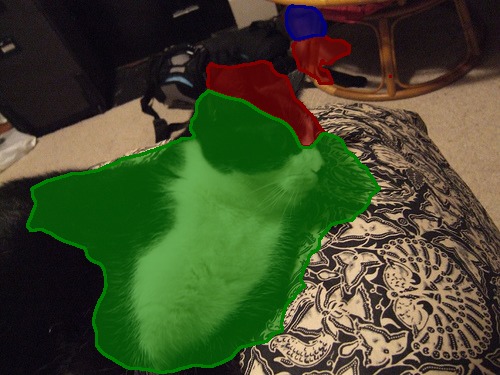}&
\includegraphics[width=0.14\linewidth]{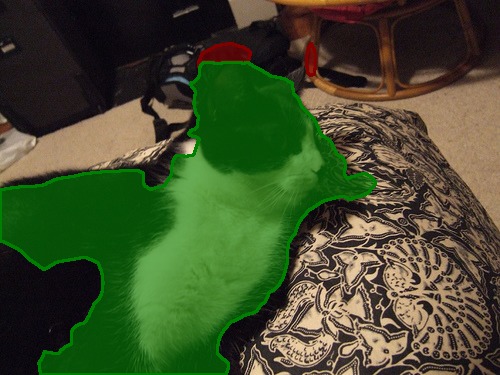}&
\includegraphics[width=0.14\linewidth]{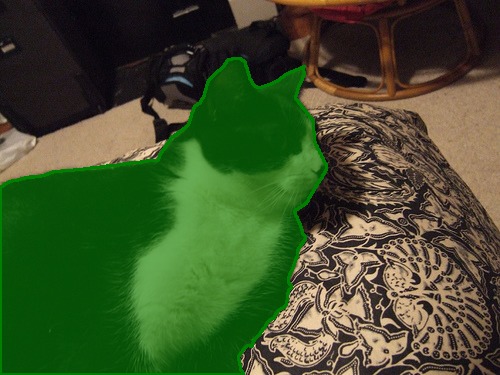}&
\includegraphics[width=0.14\linewidth]{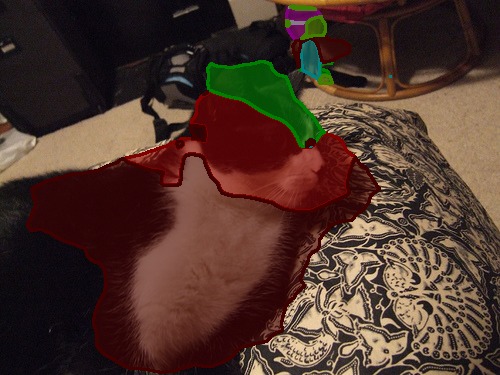}&
\includegraphics[width=0.14\linewidth]{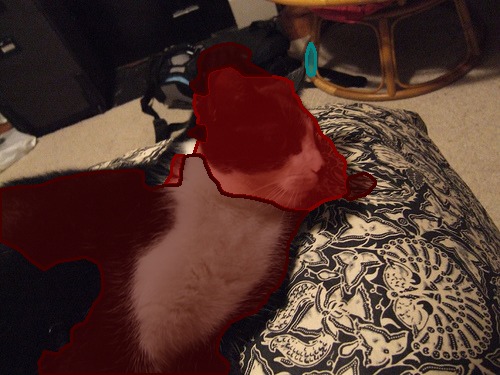}&
\includegraphics[width=0.14\linewidth]{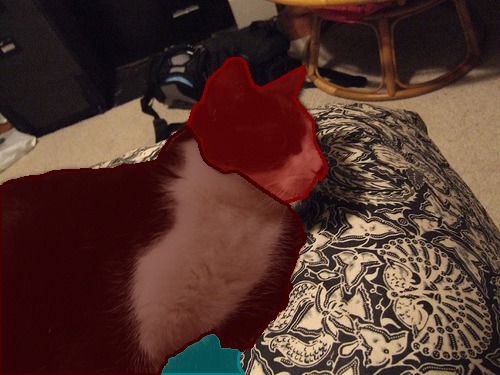}\\
\includegraphics[width=0.14\linewidth]{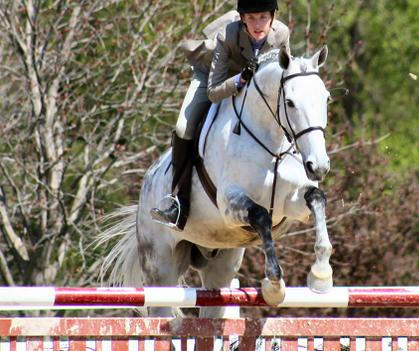}&
\includegraphics[width=0.14\linewidth]{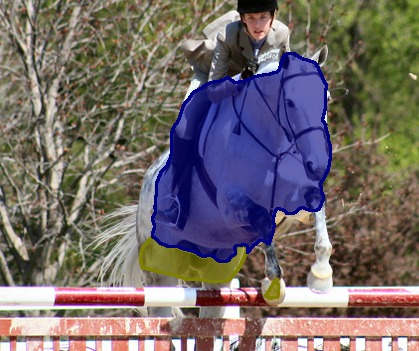}&
\includegraphics[width=0.14\linewidth]{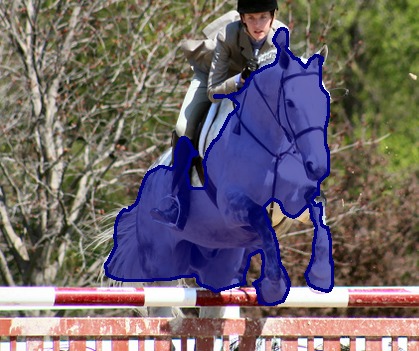}&
\includegraphics[width=0.14\linewidth]{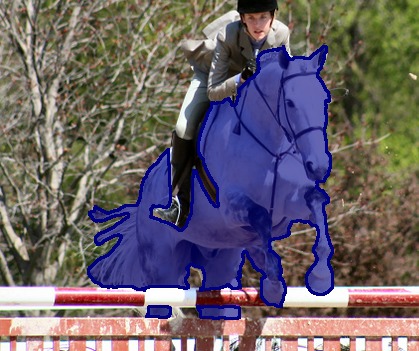}&
\includegraphics[width=0.14\linewidth]{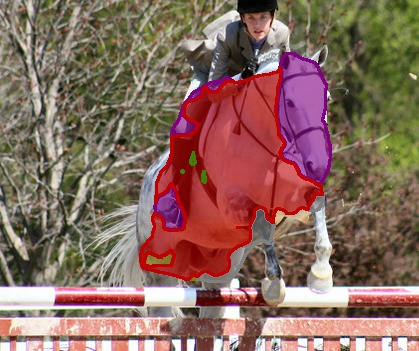}&
\includegraphics[width=0.14\linewidth]{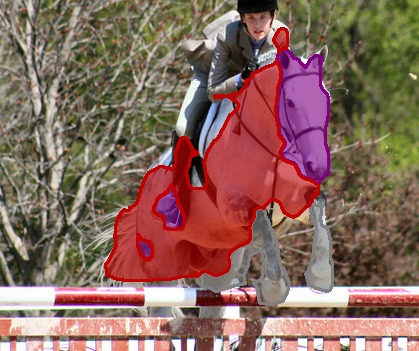}&
\includegraphics[width=0.14\linewidth]{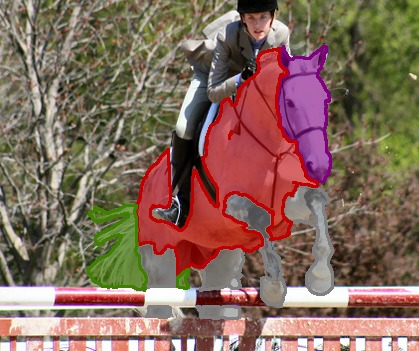}\\
\includegraphics[width=0.14\linewidth]{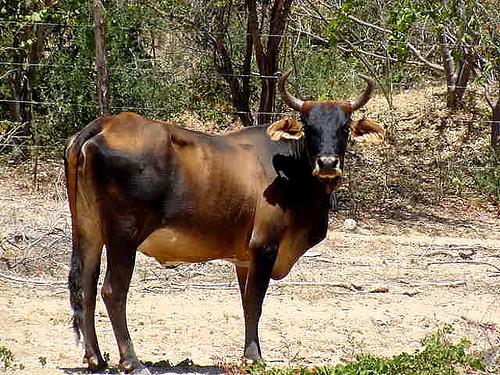}&
\includegraphics[width=0.14\linewidth]{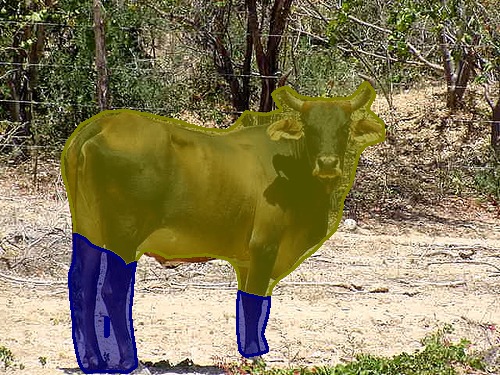}&
\includegraphics[width=0.14\linewidth]{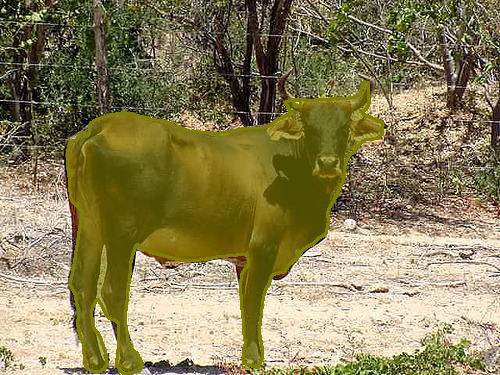}&
\includegraphics[width=0.14\linewidth]{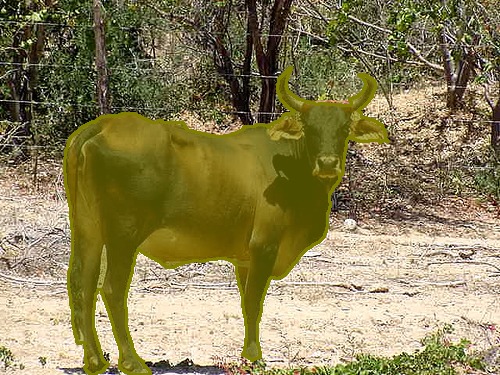}&
\includegraphics[width=0.14\linewidth]{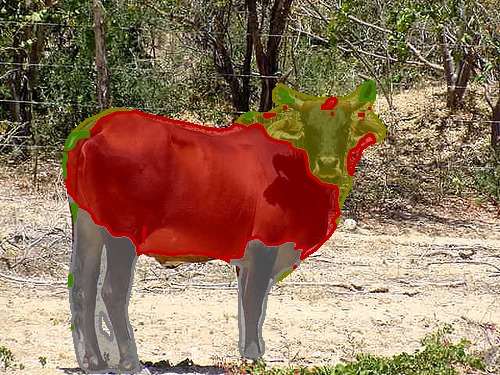}&
\includegraphics[width=0.14\linewidth]{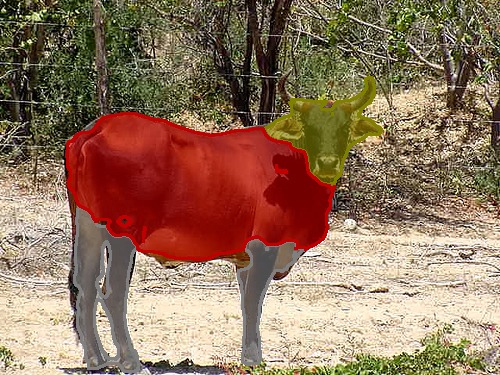}&
\includegraphics[width=0.14\linewidth]{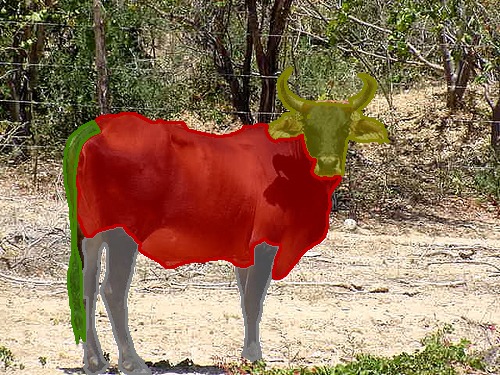}\\
Image & FCN-8s~\cite{long_shelhamer_fcn}&  Ours object & Object GT & HC~\cite{BharathCVPR2015} & Our part & Part GT.
\end{tabular}
\end{center}
   \caption{Comparison examples with FCN~\cite{long_shelhamer_fcn} for object segmentation and HC~\cite{BharathCVPR2015} for part segmentation in the Quadrupeds dataset. The  object label color map is shown at above and part label color map is shown in Fig.~\ref{fig:scp}  (Best view in color).}
\vspace{-1.\baselineskip}
\label{fig:resanimal}
\end{figure*}

Following HC, the parts of all the quadrupeds are labelled into head, body, leg and tail, from which we construct a shared SCP grammar as introduced in Sec.~\ref{subsec:grammar}. In our grammar, horse, cow and sheep share the same body and leg type, while cat and dog share others. All the animals share a tail label, and each animal has its own head label since the head is highly distinguishable~\cite{DBLP:conf/iccv/ParkhiVJZ11}. In total, 10 SCPs are used, while 20 labels are need if all the parts are treated independently..

Tab.~\ref{tab:resAnimal} shows the compared results on both object and part segmentation. As shown, in terms of object segmentation, comparing with  FCN~\cite{long_shelhamer_fcn}, our final results improves $6.2\%$, which demonstrates the advantage of the joint model. We also compare the variants of our approach with different components. ``Joint FCN(16s)'' produces the object masks directly from the two-channel FCN without FCRF inference. By including the SCP potentials, the object prediction of ``joint FCN (16s)'' already out-performs FCN, which shows that the joint object potentials have less pixel-wise confusion. In addition, as shown in the ``FCRF+FCN(16s)'', using the FCRF with the FCN object potentials without the joint convolutional layer improves over $2\%$ compared to ``FCN(16s)'', showing FCRF inference can help refine segmentation results by exploring long-range context. Moreover, with the joint potentials, the performance of our full model has another $4\%$ boost. This shows the joint FCN object potentials provide better evidence for our graphical model and is  essential to our system.   

We show several qualitative comparison examples with the FCN at left of Fig.~\ref{fig:resanimal}, in which our algorithm is able to solve the local ambiguities that FCN usually encounters. For instance, at the $4_{th}$ row, the legs of a cow are confused with horse using FCN. 
In contrast, by borrowing the object-scale evidence from the cow body and cow head, our model can correct this local confusion. In addition, as shown in the horse segments at the $3_{rd}$ row, our SCP segments are able to provide more precise object boundaries in many difficult cases, like the legs of the horse crossing the bar.  

For semantic part segmentation, we summarized the mean IOU of all parts for each object in Tab.~\ref{tab:resAnimal}. Our results are also significantly higher than the results of HC~\cite{BharathCVPR2015} (over $5\%$). As HC performs object and part segmentation sequentially, the errors in object predictions, including local confusion and inaccurate boundaries, will propagate to the parts. For example, at the $2_{nd}$ row of Fig.~\ref{fig:resanimal}, the cat and dog are confused in the object FCN prediction, which makes part of the cat head errorly labelled as dog head in HC. We solve such problems through the FCRF by considering object-scale context. In addition, thanks to our optimized image scale for both object and part, our method can capture part boundaries that are sometimes missed by the FCN and HC.

\vspace{-1.1\baselineskip}
\paragraph{PASCAL part segmentation benchmark.}
In addition to the labels from the train-validation set of~\cite{DBLP:conf/cvpr/ChenMLFUY14}, following the same hierarchical part labelling system, we additionally labelled semantic parts over the PASCAL VOC 2010 test set~\cite{Database_VOC2010} of the object segmentation task, which includes 994 images. 
With respect to the PASCAL VOC test benchmark, we are not going to release the labels and will instead launch an evaluation server for researchers to fairly compare their part segmentation results. 

We test our algorithm over images with the five quadrupeds, which include 281 images. As shown in Tab.~\ref{tab:test}, the results are consistent with that from the validation set, and our method achieves the best overall IOU outperforming the state-of-the-art with a large margin. 

\begin{table}[b]
\scriptsize
\vspace{-1.5\baselineskip}
\centering
\rowcolors{2}{}{gray!35} 
\setlength{\tabcolsep}{4pt}
\begin{tabular}{l c c c c c c c c}
\toprule[0.1 em]
\toprule[0.1 em]
&\multicolumn{8}{c}{Object segmentation accuracy}\\
& Bkg & Dog & Cat & Cow & Horse & Sheep & IOU & Pix. Acc \\
\midrule
\midrule
FCN 8s~\cite{long_shelhamer_fcn} & 94.45 &70.14 &75.45 &64.06 &64.75 &\textbf{69.06} &72.99 & 93.90\\
\midrule
Ours final  & \textbf{95.31}  & \textbf{77.44} & \textbf{80.47} & \textbf{72.13} & \textbf{76.18} & 67.96 & \textbf{78.25} & \textbf{95.26}\\
\bottomrule[0.1 em]
\end{tabular}
\rowcolors{2}{}{gray!35}
\setlength{\tabcolsep}{4pt}
\begin{tabular}{ l c c c c c c c c }
\toprule[0.1 em]
&\multicolumn{8}{c}{Semantic part segmentation accuracy}\\
& Bkg & Dog & Cat & Cow & Horse & Sheep & IOU & Pix. Acc \\
\midrule
HC~\cite{BharathCVPR2015}~~~~~~ &  94.36 &41.24 &42.42 &35.22 &45.00 &\textbf{38.86} &43.11 & 90.64\\
\midrule
Ours final &  \textbf{95.14} & \textbf{46.52} & \textbf{48.06} & \textbf{41.80} & \textbf{56.67}& 36.02& \textbf{48.16}& \textbf{92.47}\\
\bottomrule[0.1 em]
\end{tabular}
\vspace{0.1\baselineskip}
\caption{ Average precision over the part segmentation benchmark.}
\vspace{-2\baselineskip}
\label{tab:test}
\end{table}

\vspace{-0.8\baselineskip}
\section{Conclusion and discussion}
\vspace{-0.5\baselineskip}
In this paper, we proposed the framework for jointly solving object and part segmentation. Our approach follows the spirit of DPM, and leverages the advantages of both sides, yielding the state-of-the-art results. Recently, for object segmentation, there are other methods that provides additional improvement over the FCN such as adding pixel-wise Dense CRF~\cite{ChenICLR15,LINarxiv15}. Our framework is complementary to them in terms of modelling over parts and adaptive to object scale for solving local ambiguity. In addition, we tackles the part segmentation beyond the object. Possible failure cases for us would be strong appearance confusion, strong occlusion, 
where object scale or SCP segments can be misled, yielding inaccurate results (\emph{e.g}. the back leg of the dog in the first example of Fig.~\ref{fig:resanimal}). 

In the future, we will try to automatically learn the SCP, and jointly model the instance segmentation to incorporate detection, which provides better object scale and solve the remaining localization issue. 

\newpage
{\scriptsize
\bibliographystyle{ieee}
\bibliography{egbib}
}

\end{document}